\documentclass[11pt, a4paper, logo, copyright, nonumbering]{kwaipilot}
\usepackage{dblfloatfix}
\usepackage{ulem}
\usepackage{caption}
\usepackage{dramatist}
\usepackage{xspace}
\usepackage{pifont} %
\usepackage{multirow}
\usepackage{tcolorbox}
\usepackage{xltabular}
\usepackage{longtable}
\usepackage{hyperref}
\usepackage{graphicx}
\usepackage{subcaption}
\usepackage{float}
\usepackage{amsfonts}
\usepackage{amsmath}
\usepackage{amssymb}
\usepackage{lineno}
\usepackage{multirow}
\usepackage{adjustbox}
\usepackage{bbm}

\usepackage[bottom]{footmisc}

\usepackage{CJKutf8}
\usepackage{setspace}

\usepackage{dsfont}
\usepackage{array} %
\usepackage{tabularx} %
\usepackage{xcolor} %
\usepackage{tabularx}
\usepackage{booktabs}

\usepackage{lipsum}  %
\usepackage{multicol} %

\usepackage{indentfirst}
\usepackage[sort&compress,numbers]{natbib}
\newcolumntype{L}[1]{>{\raggedright\arraybackslash}p{#1}}

\usepackage{geometry}
\usepackage{makecell}
\makeatletter
\def\@BTrule[#1]{%
  \ifx\longtable\undefined
    \let\@BTswitch\@BTnormal
  \else\ifx\hline\LT@hline
    \nobreak
    \let\@BTswitch\@BLTrule
  \else
     \let\@BTswitch\@BTnormal
  \fi\fi
  \global\@thisrulewidth=#1\relax
  \ifnum\@thisruleclass=\tw@\vskip\@aboverulesep\else
  \ifnum\@lastruleclass=\z@\vskip\@aboverulesep\else
  \ifnum\@lastruleclass=\@ne\vskip\doublerulesep\fi\fi\fi
  \@BTswitch}
\makeatother

\addto\extrasenglish{
}

 {\begin{list}{}%
         {\setlength{\leftmargin}{#1}}%
         \item[]%
 }
 {\end{list}}
 
\reportnumber{001} %

\renewcommand{\today}{}

\title{\centering KAT-Coder-V2 Technical Report}

\author[*]{
KwaiKAT Team
}

\begin{abstract}
We present KAT-Coder-V2, an agentic coding model developed by the KwaiKAT team at Kuaishou. KAT-Coder-V2 adopts a \textit{Specialize-then-Unify} paradigm that decomposes agentic coding into five expert domains---SWE, WebCoding, Terminal, WebSearch, and General---each undergoing independent supervised fine-tuning and reinforcement learning, before being consolidated into a single model via on-policy distillation. We develop \textbf{KwaiEnv}, a modular infrastructure sustaining tens of thousands of concurrent sandbox instances, and scale RL training along task complexity, intent alignment, and scaffold generalization. We further propose \textbf{MCLA} for stabilizing MoE RL training and \textbf{Tree Training} for eliminating redundant computation over tree-structured trajectories with up to 6.2$\times$ speedup. KAT-Coder-V2 achieves 79.6\% on SWE-bench Verified (vs.\ Claude Opus~4.6 at 80.8\%), 88.7 on PinchBench (surpassing GLM-5 and MiniMax M2.7), ranks first across all three frontend aesthetics scenarios, and maintains strong generalist scores on Terminal-Bench Hard (46.8) and $\tau^2$-Bench (93.9). Our model is publicly available at \url{https://streamlake.com/product/kat-coder}.
\end{abstract}

\begin{document}

\maketitle

\begin{figure}[htbp]
    \centering
    \includegraphics[
        width=0.9\linewidth,
    ]{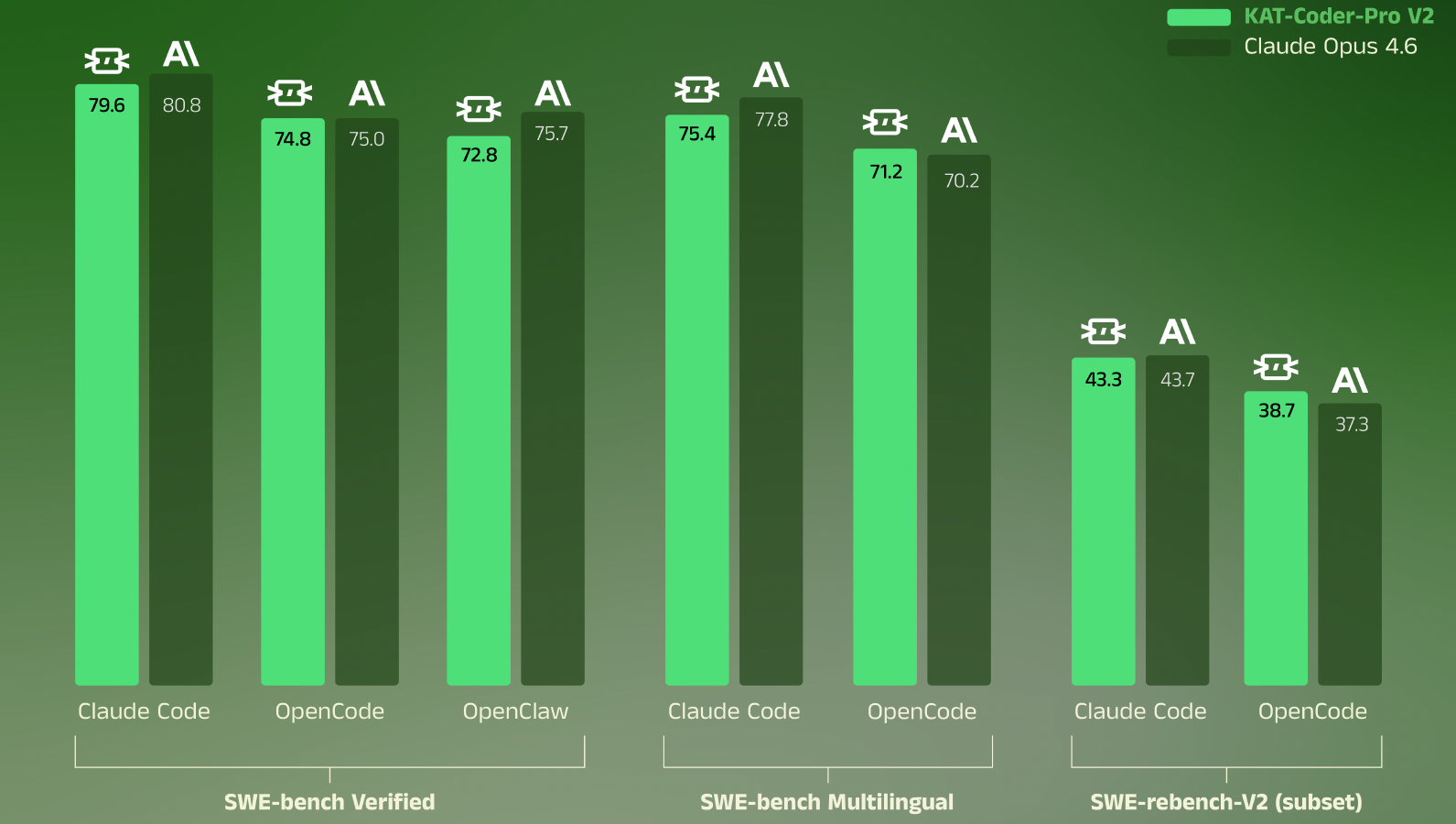}
    \caption{Results of KAT-Coder-V2 and Claude Opus 4.6 across different scaffolds on various software engineering benchmarks.}
    \label{fig:architecture}
\end{figure}

\section{Introduction}
 
Large Language Models (LLMs) are rapidly evolving from single-turn code generation toward \textbf{Agentic Coding}, the ability to autonomously plan, execute, and verify multi-step software engineering tasks within real-world development environments. Recent frontier models~\cite{claudeopus46,gemini3pro2025,zeng2026glm,team2026kimi,huang2026step,liu2025deepseek} have demonstrated impressive progress in this direction, steadily advancing the state of the art on benchmarks including SWE-bench \cite{jimenez2024swebench}, Terminal-Bench \cite{merrill2026terminal}, and $\tau^2$-Bench \cite{barres2025tau}. Unlike traditional code question-answering or mathematical reasoning, agentic coding requires models to interact with authentic code repositories, manage intricate dependency graphs, orchestrate multi-turn tool invocations, and ground their decisions in execution feedback. This interactive, long-horizon workflow demands that models' multi-step behaviors be aligned with end-to-end engineering outcomes, rather than merely optimizing for single-turn code correctness.
 
Realizing this vision presents three fundamental challenges. The first is \textbf{capability fragmentation}. SWE tasks require long-chain code editing grounded in test verification, WebCoding demands aesthetic judgment under sparse colloquial inputs, and Terminal tasks call for persistent environment state tracking. The training signals across these domains are not merely different but often conflicting, making it impractical for a single monolithic training pipeline to reach the optimum in every domain simultaneously. The second challenge is \textbf{infrastructure coupling}. Agentic RL training demands high-throughput sandbox orchestration, heterogeneous benchmark support, and seamless compatibility with a rapidly growing ecosystem of agent scaffolds such as Claude Code, OpenClaw, and OpenCode. Existing systems, however, tightly couple these concerns, making every new scaffold or dataset integration a costly engineering endeavor. The third is \textbf{scaling agentic RL}. Effectively training coding agents requires scaling along multiple dimensions simultaneously---task complexity, prompt diversity, and scaffold generalization---while coping with the MoE instability and computational redundancy introduced by tree-structured, multi-turn trajectories.
 
We introduce \textbf{KAT-Coder-V2}, a comprehensive agentic coding model developed by the KwaiKAT team at Kuaishou. Built upon KAT-Coder-V1~\cite{zhan2025kat} through continued post-training, the model follows a \textit{Specialize-then-Unify} paradigm that systematically addresses all three challenges above. We decompose the full capability spectrum into five orthogonal expert domains (SWE, WebCoding, Terminal, WebSearch, and General), each undergoing independent data construction, supervised fine-tuning, and environment-feedback reinforcement learning. The resulting domain experts are then consolidated into a single deployable model through \textbf{On-Policy Distillation (OPD)}, which combines the direct mistake-avoidance of on-policy exploration with dense, step-by-step supervision from the specialized experts, achieving lossless fusion without the exposure bias of offline imitation.
 
To tackle infrastructure coupling, we develop \textbf{KwaiEnv}, a modular infrastructure that decouples datasets, sandboxes, scaffolds, and verifiers, sustaining tens of thousands of concurrent sandbox instances. Built on this foundation, we propose an \textbf{Agentic Scaling} paradigm that systematically scales RL training along task complexity, intent alignment, and scaffold generalization, yielding over 100K diverse, high-difficulty training samples across multiple agent frameworks. To stabilize MoE RL training, we propose \textbf{MCLA} (Monte-Carlo Log-probability Averaging) for reducing log-probability variance. We further introduce \textbf{Tree Training} for eliminating redundant computation over tree-structured trajectories, achieving up to 6.2$\times$ training speedup.
 
Extensive evaluation shows that KAT-Coder-V2 closely matches Claude Opus~4.6 across scaffolds and benchmarks: 79.6\% on SWE-bench Verified (vs.\ 80.8\%), 88.7 on PinchBench (surpassing GLM-5 at 86.4 and MiniMax M2.7 at 87.1), leading scores across all three frontend aesthetics scenarios (Landing Page 59.8, Slides 57.6, Data Visualization 67.6), and strong generalist performance (Terminal-Bench Hard 46.8, $\tau^2$-Bench 93.9). These results confirm that domain-specialized training, large-scale agentic RL with systematic scaling, and unified on-policy distillation form an effective path to powerful coding agents.

\section{KwaiEnv: Infrastructure for Agentic Code Intelligence}
\label{sec:kwaienv}
\subsection{Background and Design Motivation}

As the capabilities of Large Language Models continue to evolve, Agentic Coding has emerged as a critical domain for model evaluation and Reinforcement Learning (RL) training. Unlike traditional Question-Answering (QA) or mathematical reasoning tasks, Agentic Coding—particularly Software Engineering (SWE) tasks—requires models to execute multi-step, long-chain operations within a sandbox environment equipped with authentic code repositories, dependencies, and test suites. The rollout process for these tasks involves several complex stages, including environment initialization, tool calling, state management, and result verification, far exceeding the complexity of single-turn inference scenarios.
In engineering practice, this complexity introduces the following challenges:

\begin{itemize}[leftmargin=2em]
    \item \textbf{Dataset Heterogeneity:} Diverse benchmarks (e.g., SWE-bench, SWE-bench Pro~\cite{deng2025swebenchproaiagents}) impose varying requirements on sandbox images and evaluation logic.
 
    \item \textbf{Scaffold Proliferation:} New scaffolds for Coding Agents are constantly emerging with significant differences in integration protocols; without a unified abstraction, onboarding each new agent requires redundant engineering effort.
 
    \item \textbf{High-Throughput Demands:} During the RL training phase, a massive number of rollouts must be executed concurrently, placing stringent performance requirements on sandbox scheduling and trajectory collection.
\end{itemize}

To address these challenges, we developed KwaiEnv. The core design objective is to decouple datasets, sandboxes, scaffolds, and verification logic through a modular and configurable architecture. This allows for the flexible combination of components at minimal cost, supporting the entire workflow from model evaluation to RL training.

\subsection{System Overview}

KwaiEnv provides a unified interface that supports the configurable combination of models, scaffolds, and datasets. This enables a complete closed-loop workflow encompassing model trajectory collection, rollout evaluation, and the delivery of trajectories to the RL engine for training. The system consists of five core modules, each with distinct responsibilities and high degrees of decoupling, allowing for flexible extension as needed.

\begin{figure}[tbp]
    \centering
    \includegraphics[
        width=\linewidth,
    ]{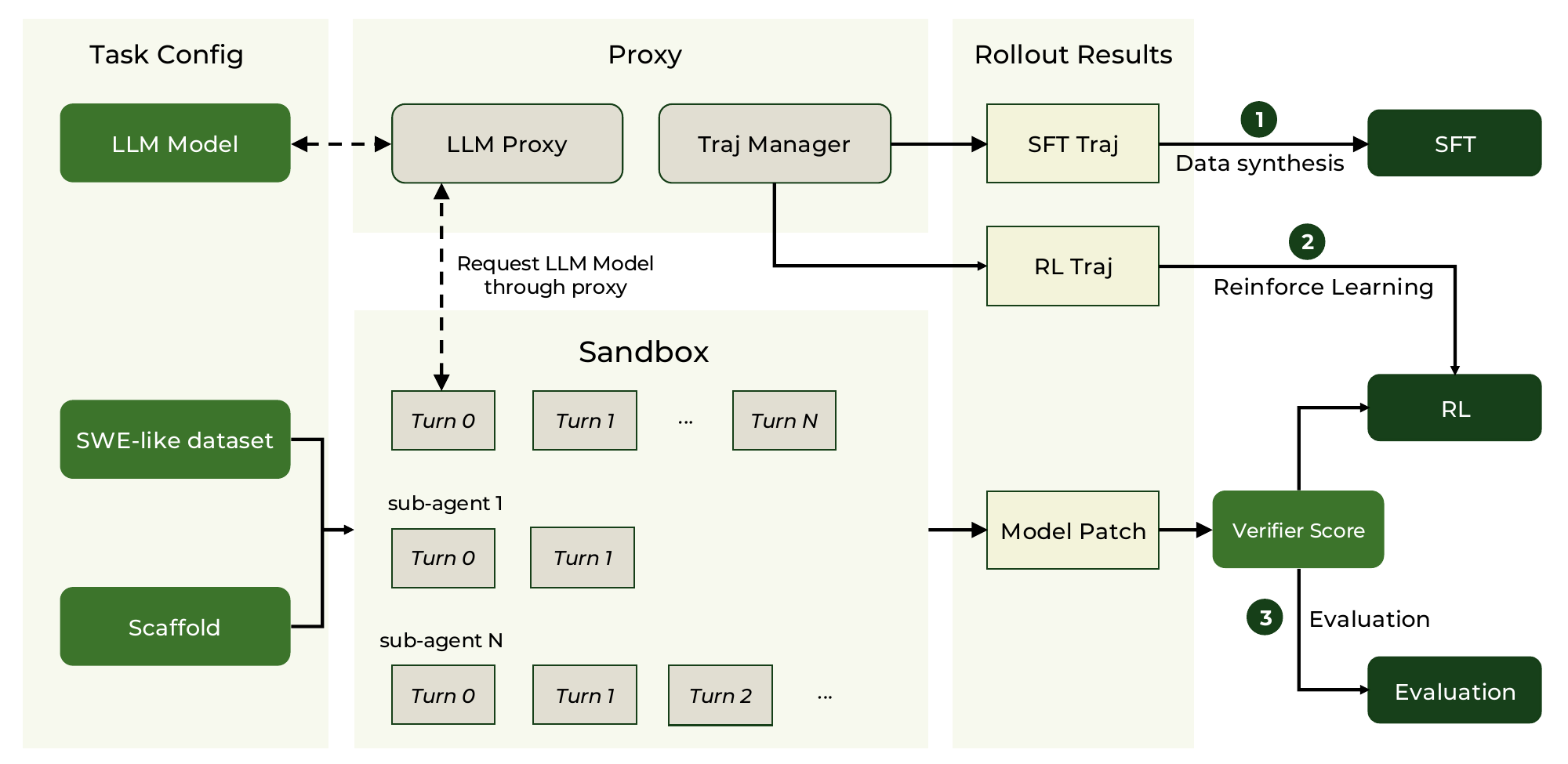}
    \caption{KwaiEnv Workflow for SWE Tasks, Supporting Key Processes Including Data Synthesis, RL, and Evaluation}
    \label{fig:kwaienv}
\end{figure}

In a typical workflow, the user specifies the dataset, target model, and scaffold via a configuration file. KwaiEnv then orchestrates the necessary remote sandboxes, deploys the scaffold onto the corresponding dataset images, and forwards model requests to the target LLM through a unified network proxy layer, recording the entire interaction trajectory. Upon completion, the Verifier scores the results, and the Trajectory Manager formats the trajectories for the RL engine. This entire pipeline operates autonomously without human intervention, significantly reducing the engineering overhead of data collection and model training, as shown in figure \ref{fig:kwaienv}.

\subsection{Core Modules}
\subsubsection{Dataset}
KwaiEnv integrates mainstream LLM benchmarks covering data analysis, code generation, SWE, web search, and general reasoning. This includes widely adopted evaluation sets such as SWE-bench~\cite{jimenez2024swebench}, LiveCodeBench~\cite{jain2024livecodebenchholisticcontaminationfree} and so on. Furthermore, the system incorporates internal proprietary training and test sets to support multi-dimensional evaluation and full-scenario RL.
The Dataset module utilizes a unified abstract interface to mask the discrepancies in task formats, image dependencies, and scoring logic across different benchmarks. New datasets can be seamlessly integrated by implementing standard methods defined by the interface.

\subsubsection{Verifier}
KwaiEnv employs differentiated verification strategies tailored to various task types, encapsulated within the Verifier module. The system supports three primary categories of verification:
\begin{itemize}[leftmargin=2em]
    \item \textbf{Deterministic Scoring:} For tasks with definitive answers (e.g., mathematical proofs, code generation), a specialized module performs precise scoring based on golden patches, execution of test cases, or standard output comparison.
 
    \item \textbf{LLM-as-Judge:} For open-ended tasks (e.g., instruction following, long-document comprehension), the system supports LLM-based evaluation and Rubric-based scoring, with configurable dimensions and weights.
 
    \item \textbf{SWE Evaluation:} For software engineering tasks, the system invokes official scoring modules to execute test suites within the sandbox and return key metrics such as pass rates.
\end{itemize}

\subsubsection{Scaffold}
KwaiEnv supports the "black-box" integration of leading Coding Agent scaffolds—including Claude Code~\footnote{https://github.com/anthropics/claude-code}, Kilo Code~\footnote{https://github.com/kilo-org/kilocode}, Cline~\footnote{https://github.com/cline/cline}, OpenClaw~\footnote{https://github.com/openclaw/openclaw}, OpenCode~\footnote{https://github.com/anomalyco/opencode}, etc —while maintaining compatibility across versions. The integration cost is minimal: since KwaiEnv proxies model requests at the network layer, any Coding Agent that calls an LLM via API can be integrated without code modifications, requiring only the configuration of API endpoints and authentication.

\subsubsection{Sandbox}
The Sandbox module is the foundational infrastructure for large-scale RL training. The system can trigger a massive number of remote sandbox instances within seconds. Each sandbox runs in an isolated container environment, mounted with dataset-specific images. KwaiEnv manages the entire lifecycle—creation, task assignment, monitoring, and reclamation—making the process transparent to upper-layer modules. The system can support tens of thousands of concurrent sandboxes, providing the high throughput required for rapid RL rollout acquisition.

\subsubsection{Trajectory Manager}
Acting as the bridge between KwaiEnv and the RL engine, the Trajectory Manager handles trajectory collection, formatting, and output. It intercepts all LLM requests via the network proxy, recording comprehensive metadata including I/O content, tool-call sequences, token usage, and timestamps.
For RL training, the module can assemble, reorder, and truncate raw trajectories to meet the input specifications of various algorithms.

\subsection{Decoupling and Scalability}
KwaiEnv adheres to the principle of Separation of Concerns. The five core modules communicate through standardized interfaces, allowing independent iteration of any module. This design yields several key benefits:
\begin{itemize}[leftmargin=2em]
    \item \textbf{Data Scalability:} Scaling training data requires only the implementation of a unified data interface, without impacting sandboxes or scaffolds.
 
    \item \textbf{Scaffold Scalability:} New Coding Agents can be onboarded by simply configuring container commands and API endpoints.
 
    \item \textbf{Evaluation Agility:} The evaluation and training pipelines share the same infrastructure, ensuring high consistency and short iteration cycles.
 
    \item \textbf{Algorithmic Adaptability:} The formatting logic is decoupled from RL algorithms; new algorithms can be supported by simply registering new trajectory formatting rules.
\end{itemize}

\section{Post-Training Methodology}
\label{sec:posttraining}
\subsection{Training Pipeline Overview}
 
KAT-Coder-V2 is built upon KAT-Coder-V1~\cite{zhan2025kat} through continued post-training, following a \textit{specialize-then-unify} paradigm. We decompose the capability spectrum of agentic coding into five orthogonal expert domains---SWE (software engineering repair and development), WebCoding (frontend generation and aesthetics), Terminal (command-line reasoning), WebSearch (online search and information synthesis), and General (general-purpose code intelligence)---each of which undergoes independent data construction and specialized training.
 
The overall pipeline consists of three stages:
\begin{itemize}
    \item \textbf{Supervised Fine-Tuning:} For each expert domain, we leverage KwaiEnv's trajectory collection capabilities and domain-specific data synthesis pipelines to construct large-scale, high-quality training data, producing a dedicated expert model per domain.
    \item \textbf{Reinforcement Learning:} Using the sandbox environments and verifier infrastructure provided by KwaiEnv, we apply environment-feedback-based reinforcement learning to further improve decision quality in multi-turn interactions and long-horizon tasks.
    \item \textbf{On-Policy Distillation:} The capabilities of multiple domain experts are consolidated into a unified KAT-Coder-V2 through on-policy distillation, achieving single-model deployment while retaining expert-level performance across all domains.
\end{itemize}
 
The following subsections detail the data construction and training methodology for each expert domain.

\subsection{Supervised Fine-Tuning}
 
We train five domain experts via supervised fine-tuning, each targeting a distinct capability required for agentic coding. Table~\ref{tab:expert_overview} summarizes the data sources, scale, and key methodological innovations of each expert. The remainder of this section details the unique technical contributions within each domain.
 
\begin{table}[tbp]
\centering
\caption{Overview of the five expert domains in the SFT stage.}
\label{tab:expert_overview}
\small
\renewcommand{\arraystretch}{1.3}
\begin{tabularx}{\linewidth}{l l X}
\toprule
\textbf{Expert} & \textbf{Scenario} & \textbf{Key Methodology} \\
\midrule
SWE       & Issue resolution   & Issue-PR pairing with merge-status supervision; AutoBuilder for verifiable task synthesis (F2P+P2P); Code Comprehension trajectory synthesis \\
WebCoding & UI generation      & Tri-Perspective label system; Prompt rewriting (designer $\rightarrow$ ordinary user); designer-panel evaluation \\
Terminal  & CLI reasoning      & Cross-format SWE$\rightarrow$Terminal conversion; multi-agent synthesis; Docker-based verification \\
WebSearch & Agentic search     & KG construction from search trajectories; Pass@8 filtering; rejection sampling fine-tuning \\
General   & Instr.\ / QA / Code-Math & Compositional constraint training; long-conversation samples; online-judge verification \\
\bottomrule
\end{tabularx}
\end{table}

\subsubsection{SWE Expert: Autonomous Issue Resolution}
 
The SWE Expert targets real-world software engineering scenarios, training the model to autonomously perform end-to-end tasks---codebase comprehension, fault localization, and code repair---starting from an issue description. Data construction revolves around three complementary pipelines: \textbf{Issue-PR}, which supplies large-scale real-world engineering repair corpora, \textbf{AutoBuilder}, which generates verifiable interactive training tasks, and \textbf{Code Comprehension}, which produces interactive code understanding trajectories grounded in real-world repositories.
 
\begin{figure}[tbp]
    \centering
    \includegraphics[
        width=\linewidth,
    ]{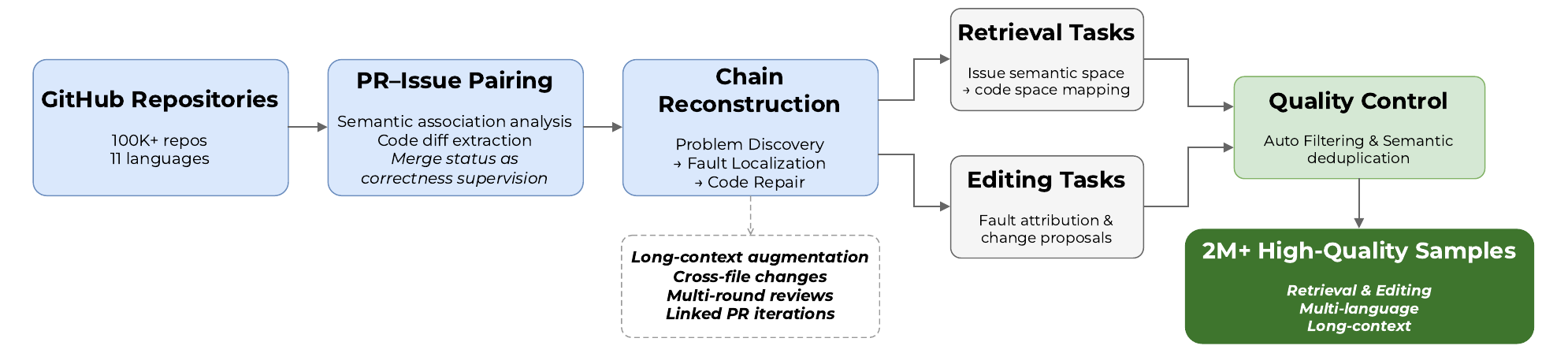}
    \caption{Overview of the Issue-PR data construction pipeline. Merged PRs serve as anchor points for bidirectional Issue--PR mapping and code diff extraction, with merge status providing natural correctness supervision. The reconstructed engineering chains are decomposed into retrieval and editing tasks, then filtered through multi-stage quality control to yield over 2M training samples.}
    \label{fig:issue_pr_pipeline}
\end{figure}

\paragraph{Issue-PR Pipeline.}
We extract paired data of merged Pull Requests and their associated Issues from hundreds of thousands of GitHub open-source repositories, covering 11 mainstream programming languages (illustrated in Figure~\ref{fig:issue_pr_pipeline}). Using merged PRs as anchor points, we establish bidirectional Issue-PR mappings through semantic association analysis. Specifically, for each merged PR $p$, we compute a relevance score
\begin{equation}
s(i, p) = \cos\bigl(\mathbf{e}_i,\, \mathbf{e}_p\bigr)
\end{equation}
between the Issue embedding $\mathbf{e}_i$ and the PR embedding $\mathbf{e}_p$, retaining pairs with $s(i, p) > \tau$ to establish bidirectional mappings. We then extract pre- and post-merge code state differences (diffs), and reconstruct the complete \textit{problem discovery $\rightarrow$ fault localization $\rightarrow$ code repair} chain.
 
Building upon this chain, we construct two complementary training paradigms. \textit{Retrieval tasks} guide the model to perform precise mapping from the Issue semantic space to the code space---given an Issue description, the model must locate relevant files and functions within a large-scale codebase. \textit{Editing tasks} require the model to produce complete repairs that integrate fault attribution with change proposals, forming an end-to-end capability loop. Along the long-context dimension, we exploit the inherent long-range dependency characteristics of Issue-PR data (cross-file changes, multi-round reviews, and linked PR iterations) by aggregating highly correlated engineering fragments into long-sequence samples, strengthening the model's ability to associate information across large-scale codebases.
 
Regarding data quality, the PR merge status serves as a natural correctness supervision signal. On top of this, we filter out auto-generated artifacts and non-essential dependency changes, perform semantic-level deduplication of repetitive repair patterns, and ultimately curate over 2M high-quality samples.
 
\begin{figure}[tbp]
    \centering
    \includegraphics[
        width=\linewidth,
    ]{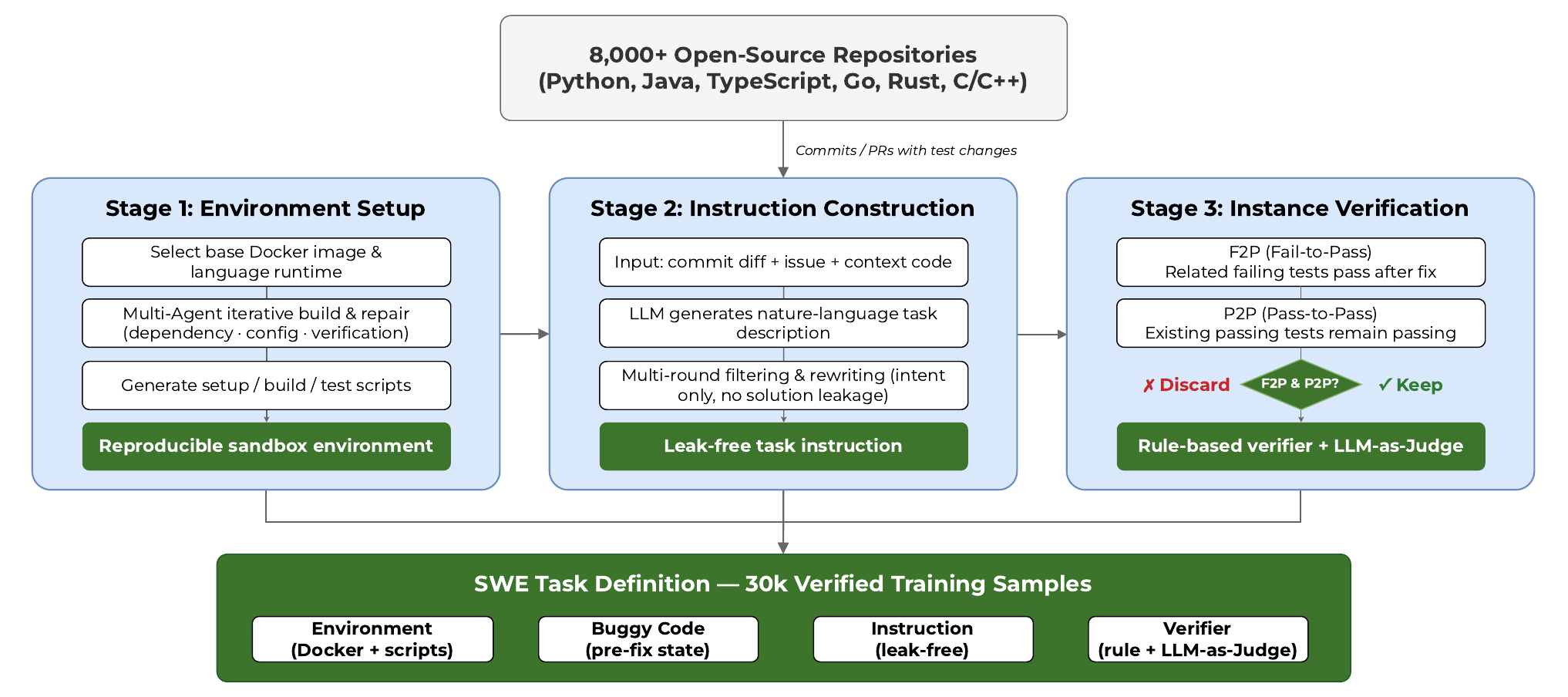}
    \caption{Overview of the AutoBuilder pipeline for automated SWE task synthesis from open-source repositories.}
    \label{fig:swe_pipeline}
\end{figure}

\paragraph{AutoBuilder Pipeline.}
Static code data lacks environment interaction information and is insufficient for training the long-horizon reasoning capabilities required in agentic scenarios. To address this, we design an automated task synthesis pipeline (illustrated in Figure~\ref{fig:swe_pipeline}) that automatically constructs verifiable software engineering tasks from real-world repositories, comprising three stages:
 
\textit{Environment Setup.}
We select active repositories with well-configured CI from GitHub and extract commit/PR instances that contain unit test changes. For each instance, we employ multi-agent collaboration to automatically construct an isolated sandbox: a Dependency Resolution Agent, an Environment Configuration Agent, and a Build Verification Agent are respectively responsible for dependency installation, compilation configuration, and test execution. These agents iteratively set up and repair the environment based on the repository's own Dockerfile, dependency manifests, and CI scripts, until the code compiles and tests are executable.
 
\textit{Instruction Construction.}
Taking the commit diff, associated Issue, and surrounding code context as input, we use an LLM to automatically generate user instructions. The key constraint is that instructions must describe only the requirement intent without leaking implementation details. Multi-round filtering ensures clarity and open-endedness, closely approximating the way real users pose questions.
 
\textit{Instance Verification.}
Let $\mathcal{T}_{\text{fail}}$ and $\mathcal{T}_{\text{pass}}$ denote the sets of originally failing and passing tests, respectively. An instance with repaired code $\hat{c}$ is retained if and only if it satisfies both criteria:
\begin{equation}
\underbrace{\forall\, t \in \mathcal{T}_{\text{fail}}:\ t(\hat{c}) = \text{Pass}}_{\text{Fail-to-Pass (F2P)}}
\quad \wedge \quad
\underbrace{\forall\, t \in \mathcal{T}_{\text{pass}}:\ t(\hat{c}) = \text{Pass}}_{\text{Pass-to-Pass (P2P)}}
\end{equation}
F2P confirms the repair is effective by requiring all previously failing tests to pass, while P2P rules out regression defects by ensuring all previously passing tests remain unaffected. Only instances satisfying both conditions are retained.
 
Through this pipeline, we produce 30k verified training samples from over 8,000 open-source repositories spanning mainstream languages including Python, Java, TypeScript, Go, Rust, and C/C++, covering typical task types such as bug fixing, feature development, and code refactoring. Each sample is defined by a complete quadruple: a reproducible environment (Docker image + build scripts), buggy-state code, a leak-free task instruction, and a dual verification mechanism combining a rule-based verifier with multi-dimensional GRM scoring.
 
\paragraph{Code Comprehension Pipeline.}
While the Issue-PR and AutoBuilder pipelines focus on code \textit{editing} capabilities, agentic SWE equally demands deep code \textit{comprehension}---the ability to navigate, understand, and reason about large-scale codebases. To train this complementary skill, we design a seven-stage trajectory synthesis pipeline that produces interactive code understanding data grounded in real-world repositories.

The pipeline begins with large-scale repository discovery: we crawl high-star GitHub repositories via segmented search (partitioning by star ranges to bypass the API's 1,000-result limit) and apply a six-dimensional quality filter---covering naming patterns, description keywords, language composition ($\geq$50\% primary-language code), contributor count ($\geq$10), PR/Issue activity ($\geq$50 each), and the presence of source code or build configuration files---to retain only repositories genuinely suitable for code comprehension tasks. For each qualifying repository, we retrieve structured project documentation via DeepWiki and pin the corresponding commit hash to ensure version consistency with the subsequent sandbox environment.

Next, we construct isolated Docker environments per repository (based on the pinned commit) and synthesize code comprehension queries using an LLM. The query synthesis is guided by a controlled design covering six question types (overview, code locating, implementation walkthrough, call-chain tracing, enhancement planning, and code review) across four difficulty levels, with balanced Chinese--English bilingual generation, yielding approximately eight queries per repository.

Trajectory synthesis is then performed by deploying a Claude Code Agent inside each Docker container, where the agent autonomously explores the codebase using its full toolset (file reading, grep search, bash execution) to answer the generated queries, with a maximum of 150 interaction turns per task. The resulting raw trajectories are converted from Anthropic format to OpenAI-compatible training format.
 
\subsubsection{WebCoding Expert: Aesthetic-Aware UI Generation}
 
The WebCoding Expert targets automatic generation of frontend pages (HTML/CSS/JS) with commercial-grade visual quality from natural language input, focusing on Landing Pages, Presentations, and Data Visualizations. The core challenge is that real users predominantly provide colloquial, short-form inputs (e.g., ``make it cool and street-style''), while general-purpose models suffer from \textit{aesthetic collapse} under such low-information-density inputs, regressing to conservative defaults (blue-white palettes, single-column grids).
 
\paragraph{Tri-Perspective Label System.}
We propose a label system that maintains three aligned views for each design specification: \textit{user perception} $\rightarrow$ \textit{design rationale} $\rightarrow$ \textit{technical implementation}. Formally, the system defines a structured mapping $\mathcal{L}: V_{\text{user}} \rightarrow V_{\text{design}} \rightarrow V_{\text{impl}}$, decomposed into seven hierarchical levels $\{L_k\}_{k=1}^{7}$ (L1 style guidance $\rightarrow$ L2--L4 global visual/animation/typography norms $\rightarrow$ L5--L7 module-level specifications/technical implementation/asset manifests). Colloquial user inputs typically cover only $L_1$; the model autoregressively infers the remaining levels:
\begin{equation}
\hat{L}_{k} = f_\theta\bigl(L_1,\, \hat{L}_2, \ldots, \hat{L}_{k-1}\bigr), \quad k = 2, \ldots, 7
\end{equation}
transforming the generation from a black box into a traceable, structured derivation.
 
\paragraph{Data Synthesis and Prompt Rewriting.}
Data construction proceeds in four stages: high-quality design screenshot collection, reverse-engineered structured prompts, seed HTML generation with designer screening, and large-scale training data derivation via a Teacher Model. To bridge the distributional gap between verbose structured prompts and real user inputs, we adopt a \textit{Prompt Rewriting} strategy: for each HTML, we construct three semantically equivalent prompt variants---a \textit{designer-annotated} version ($>$1000 words), a \textit{professional-user} version (200--300 words), and an \textit{ordinary-user} version ($\sim$50 words). This spectrum enables consistent visual quality across varying input granularities.
 
\paragraph{Aesthetic Evaluation.}
A critical distinction in frontend generation is between \textit{code fidelity}---whether the generated HTML/CSS renders correctly without errors---and \textit{aesthetic fidelity}---whether the rendered page achieves professional-grade visual quality as judged by trained designers. Code fidelity is a necessary but insufficient condition for aesthetic fidelity: a page that renders without errors can still score anywhere from poor to excellent on aesthetic quality. Existing benchmarks (e.g., WebArena~\cite{zhou2023webarena}, Design2Code~\cite{si2025design2code}) predominantly measure code fidelity or pixel-level similarity against a reference design, leaving a systematic gap in evaluating aesthetic quality---particularly in the \textit{reference-free} Text-to-UI setting where no ground-truth design exists.

To address this gap, we establish the first systematic reference-free aesthetic evaluation benchmark for Text-to-UI generation. All test prompts are drawn exclusively from colloquial, ordinary-user inputs (e.g., ``make it cool and street-style''), directly testing the model's ability to infer complete aesthetic decisions from low-information-density descriptions.

For \textbf{Landing Pages}, we decompose aesthetic fidelity into 10 independent dimensions spanning four layers:
\begin{itemize}
    \item \textit{Structural layer}: \textbf{Layout} (spatial rhythm, alignment consistency, visual hierarchy) and \textbf{Typography} (font-size hierarchy, weight contrast, line spacing);
    \item \textit{Visual layer}: \textbf{Color} (primary/secondary/accent color system coherence), \textbf{Font} (typeface selection and scene appropriateness), \textbf{Image} (style consistency and thematic relevance), \textbf{Background} (section differentiation and gradient quality), and \textbf{Elements} (icon consistency, SVG quality, logo fidelity);
    \item \textit{Component layer}: \textbf{Components} (button/card/form style consistency and primary--secondary visual distinction);
    \item \textit{Dynamic layer}: \textbf{Interaction} (hover/active/focus feedback richness) and \textbf{Animation} (entry sequence design, timing, and visual appeal).
\end{itemize}

For \textbf{Presentations (Slides)}, we adopt a streamlined 5-dimension evaluation comprising Layout, Typography, Color, Image, and Elements. The remaining five dimensions are removed because: (i)~font selection in slides conventionally defaults to system fonts for cross-platform compatibility; (ii)~uniform backgrounds across slides are standard practice; (iii)~slide components are structurally simpler than Landing Page counterparts; and (iv)~slide transitions and interactions are baseline expectations rather than aesthetic differentiators.

Each dimension is scored on a 0--5 scale with precisely anchored rubrics: 0 denotes complete absence, 1 indicates rendering failure, and 5 requires flawless execution with notable design excellence. All evaluations are conducted by a calibrated professional UI/UX designer panel under standardized conditions (Chrome, 1920$\times$1080 viewport, full interactive review including scroll, hover, and click), ensuring that dynamic dimensions such as interaction and animation are properly assessed rather than judged from static screenshots alone.

\subsubsection{Terminal Expert: Interactive Command-Line Reasoning}
 
The Terminal Expert targets complex tasks in real terminal environments---system configuration, experiment reproduction, DevOps operations, and general software engineering---requiring broad domain knowledge and autonomous decision-making. Each training sample contains a task instruction, a reference solution, an automated test script, and a reproducible Docker environment. We construct data from four complementary sources:
 
\begin{itemize}
    \item \textbf{Expert-annotated data}: domain experts manually author tasks across 12 technical domains (DevOps, data science, security, computational biology, etc.), screened for sufficient difficulty via mainstream model evaluation.
    \item \textbf{Multi-agent synthetic data}: over 20 parallel agent instances automatically produce verifiable tasks including descriptions, Docker environments, and test scripts.
    \item \textbf{Cross-format adaptation}: SWE-format tasks are converted into Terminal format via AutoBuilder, yielding 100K+ verifiable tasks across 10 programming languages.
    \item \textbf{Open-source integration}: established datasets including CLI-Gym~\cite{lin2026cligym}, and TermiGen~\cite{zhu2026termigen}, covering 420+ unique CLI tools across 11 task categories.
\end{itemize}

\subsubsection{WebSearch Expert: Agentic Search}
 
The WebSearch Expert trains the model to answer complex questions by actively invoking search tools and performing multi-hop inference. We construct over 100K training samples through a pipeline centered on search-trajectory-based knowledge graph construction and multi-stage filtering.
 
\paragraph{Knowledge Graph Construction from Search Trajectories.}
Within a single search trajectory, the web pages visited sequentially form a naturally coherent document set. We exploit this coherence to construct knowledge graphs: named entities are extracted and linked via co-occurrence relations across pages, forming bridging nodes. Multi-hop subgraphs are sampled along paths of controllable depth, and an LLM generates QA pairs by masking key entities---ensuring questions cannot be answered by parametric memory alone.
 
\paragraph{Filtering and Rejection Sampling.}
Raw data undergoes two rounds of filtering. First, samples answerable without tools are removed. Second, each sample is independently sampled $K{=}8$ times, yielding an empirical success rate:
\begin{equation}
\hat{r} = \frac{1}{K}\sum_{j=1}^{K}\mathbbm{1}[a_j = a^*]
\end{equation}
where $a_j$ is the $j$-th sampled answer and $a^*$ the ground truth. Trivially easy ($\hat{r}{=}1$) and intractable ($\hat{r}{=}0$) samples are discarded, retaining only the intermediate band that maximizes the effective gradient signal for policy optimization. Rejection sampling fine-tuning then selects positive trajectories satisfying three criteria: correct final answer, no failed tool calls, and no duplicate queries.

\subsubsection{General Expert: Instruction Following and Code-Math Reasoning}
 
The General Expert maintains the model's core competitiveness across general-purpose scenarios, covering three directions: \textit{instruction following} (format, content, and compositional constraints with fine-grained violation penalties), \textit{general QA} (open-domain knowledge, multi-turn dialogue with long-conversation samples involving topic shifts and cross-turn dependencies), and \textit{code-math reasoning} (competition-level programming and mathematical problem sets from elementary to advanced levels, verified through online judge systems). This expert ensures fundamental capabilities are retained while domain-specific experts are strengthened.


\subsection{Reinforcement Learning}

\subsubsection{Agentic RL}

\paragraph{Agentic Scaling.}

While SFT establishes a model's foundational instruction-following and code-generation capabilities, Reinforcement Learning is crucial for advancing its exploration and reasoning in complex, long-horizon tasks. However, conventional RL datasets---typically derived from simple question-answering pairs or static environments---fail to capture the inherent variability and complexity of real-world agentic scenarios.

To bridge this gap, we propose an RL data synthesis paradigm termed \textbf{Agentic Scaling}. Leveraging a foundational task pool curated by our internal \textit{Autobuilder} system, we systematically scale the training data across three critical dimensions: \textit{Task Complexity}, \textit{Intent Alignment}, and \textit{Scaffold Generalization}. This pipeline yields a large-scale, high-quality RL dataset comprising over 100,000 diverse samples.

Effective policy optimization requires training on tasks poised near the model's capability frontier. Using the \textit{Autobuilder} pool, we employ a state-of-the-art closed-source model acting as both \textbf{Teacher} and \textbf{Judge} to generate and robustly verify trajectories within a secure sandbox. We explicitly filter out easily solvable tasks, retaining only challenging instances that necessitate extensive reflection or iterative refinement, even for the frontier teacher model. These verified, high-difficulty trajectories provide the critical learning signals necessary to unlock deeper reasoning during RL.

A primary challenge in real-world deployment is a distinct \textbf{Sim-to-Real Gap}: whereas training data typically features well-structured, expert-crafted prompts, end-users frequently provide incomplete or ambiguous instructions. To enhance robustness against this discrepancy, we apply semantic augmentation to task descriptions, ensuring that \textbf{each real-world code commit maps to a diverse set of prompts}. Using LLMs, we rewrite standardized task specifications into a spectrum of variants---ranging from detailed expert instructions to colloquial, underspecified queries. This ``one-commit-to-multiple-prompts'' strategy compels the model to accurately infer the underlying engineering intent from noisy, realistic user inputs.

Furthermore, to prevent overfitting to any single agent framework, we treat the \textbf{scaffold} itself as an independent variable during data synthesis. We generate trajectories using \textbf{black-box scaffolds} (Claude Code, OpenCode, Kilo Code, etc.), which operate in highly abstracted environments and emphasize final task outcomes, alongside \textbf{white-box} variants (SWEagent~\footnote{https://github.com/swe-agent/swe-agent}). Training the model across multiple scaffolds for the identical task fosters scaffold-agnostic, highly transferable problem-solving behaviors.

Finally, we generalize the RL format into a unified \textbf{5-tuple} representation:
\begin{equation}
\mathcal{D}_{\text{RL}} = \left\{ \langle \mathcal{E}, \mathcal{T}_{\text{tools}}, \mathcal{S}_{\text{agent}}, \mathcal{I}_{\text{task}}, \mathcal{V}_{\text{verifier}} \rangle \right\}
\end{equation}
where $\mathcal{E}$ denotes the execution environment, $\mathcal{T}_{\text{tools}}$ the available toolset, $\mathcal{S}_{\text{agent}}$ the specific scaffold and system prompt, $\mathcal{I}_{\text{task}}$ the task instruction, and $\mathcal{V}_{\text{verifier}}$ the verification and reward signals. This comprehensive formulation captures the rich supervision required for interaction-heavy coding tasks, establishing a robust data foundation for scaling agentic RL.

\paragraph{Modified Turn-level Policy Optimization.}

While Group Relative Policy Optimization (GRPO~\cite{shao2024deepseekmath}) efficiently eliminates the value model, its token-level importance sampling can introduce high variance in long-horizon agent scenarios like Software Engineering (SWE). Conversely, Group Sequence-level Policy Optimization (GSPO)~\cite{zheng2025group} improves stability by aggregating probabilities across the entire trajectory. However, applying a single sequence-level ratio and advantage makes temporal credit assignment challenging in multi-turn environments, as it obscures which specific turn (e.g., Turn 1 vs. Turn 5) led to the ultimate outcome.

To balance training stability and precise credit assignment, we introduce a \textbf{turn-level adaptation} of GSPO. We operate at the granularity of an \textit{interaction turn} by partitioning the full generated sequence $y$ into $N$ discrete turns. For each turn $n$ containing a subset of tokens $\mathcal{T}_n$, we compute an independent importance ratio:
\begin{equation}
r_{turn}^{(n)}(\theta) = \prod_{i \in \mathcal{T}_n} \frac{\pi_\theta(y_i | x, y_{<i})}{\pi_{\theta_{old}}(y_i | x, y_{<i})}
\end{equation}
The clipped surrogate objective over a group of trajectories $\tau$ is then formulated as:
\begin{equation}
\mathcal{L}^{Turn}(\theta) = \mathbb{E}_{\tau \sim \pi_{\theta_{old}}} \left[ \frac{1}{N} \sum_{n=1}^N \min \left( r_{turn}^{(n)}(\theta) A_n, \text{clip}\left(r_{turn}^{(n)}(\theta), 1-\epsilon, 1+\epsilon\right) A_n \right) \right]
\end{equation}
where $A_n$ is the group-level advantage. 

By evaluating the probability shift of entire action blocks, this formulation deeply aligns with the Markov Decision Process (MDP) of LLM Agents. It preserves the variance reduction benefits of sequence-level optimization while ensuring fine-grained credit assignment. Furthermore, dynamically defining turn boundaries based on scaffold-specific markers seamlessly accommodates our multi-scaffold data, significantly accelerating convergence and enhancing the model's self-correction capabilities in long-step debugging.

\paragraph{Monte-Carlo Logprob Averaging.} 
RL training of Mixture-of-Experts (MoE) models is widely known to be unstable, often attributed to policy mismatch between rollout and training phases. In addition to this, we identify another key factor: the high variance of trajectory log-probability expectation estimates, which leads to unstable gradient directions during optimization. Specifically, policy gradient methods rely on Monte Carlo estimation of the form:
\begin{equation}
\nabla J(\theta) = \mathbb{E}_{a \sim \pi_{\text{rollout}}}\left[ 
R(a)\, \frac{\pi_{\text{train}}(a)}{\pi_{\text{rollout}}(a)} \nabla \log \pi_{\text{train}}(a) 
\right]
\end{equation}
where the importance weight depends on estimated log-probabilities. In practice, due to the inherent stochasticity of MoE architectures (e.g., stochastic expert routing, capacity dropping, or numerical variance), the estimated policy log-probability is noisy:
\begin{equation}
\log \pi(a) = \log \pi^*(a) + \epsilon
\end{equation}
This noise induces high variance in the importance weights:
\begin{equation}
w(a) = \exp\left(\log \pi_{\text{rollout}}(a) - \log \pi_{\text{train}}(a)\right)
\end{equation}
As a result, the variance of the estimator, $\mathrm{Var}\left[R(a)\, w(a)\, \nabla \log \pi_{\text{train}}(a) \right]$, can become excessively large, leading to unstable gradient directions.

To address this, we adopt a simple yet effective variance reduction strategy termed \textbf{MCLA} (Monte-Carlo Log-probability Averaging). During training, the forward pass for each trajectory is prefetched (prefilled) $K$ times, and the corresponding log-probabilities are averaged:
\begin{equation}
\bar{\log \pi}(a) = \frac{1}{K} \sum_{k=1}^{K} \log \pi^{(k)}(a), \quad K=8
\end{equation}
thereby significantly reducing the variance of the trajectory-level estimator. In addition, we combine MCLA with IcePop (which suppresses training–inference misalignment in RL training by clipping excessive-discrepancy tokens), aligning routing decisions between rollout and training, and further reducing system-level mismatch. These two components are strictly complementary: log-probability averaging reduces estimator variance, while IcePop mitigates distributional inconsistency. Empirically, this synergy results in highly stable training, faster convergence, and superior final performance.

\subsubsection{Agentic Engineering}

\paragraph{RL Framework.} 

To robustly support our policy optimization algorithms and the massive scale of agentic data, we developed \textbf{KRL (Kwai RL)}, a highly optimized reinforcement learning framework built around two core system innovations. First, \textbf{Tree Training} addresses the severe computational bottleneck of group sampling by eliminating the redundant calculation of shared prefixes across trajectory branches, achieving an approximate 6$\times$ acceleration during training. Second, KRL is engineered for \textbf{high-efficiency, large-scale sandbox environment training}. To handle the complex and asynchronous interactions between the policy model and diverse execution scaffolds, we integrated Cache-Aware intelligent scheduling to maximize KV Cache hit rates and Dynamic Streaming for fine-grained pipeline orchestration. By seamlessly interleaving the generation (Rollout) and weight update (Training) phases across massive sandbox instances, this architecture reduces the overall unit sample cost by 2.8$\times$, providing an essential, cost-effective infrastructural foundation for agentic scaling.

\paragraph{Tree Training.}

\begin{figure}[tbp]
    \centering
    \includegraphics[
        width=1.0\linewidth,
    ]{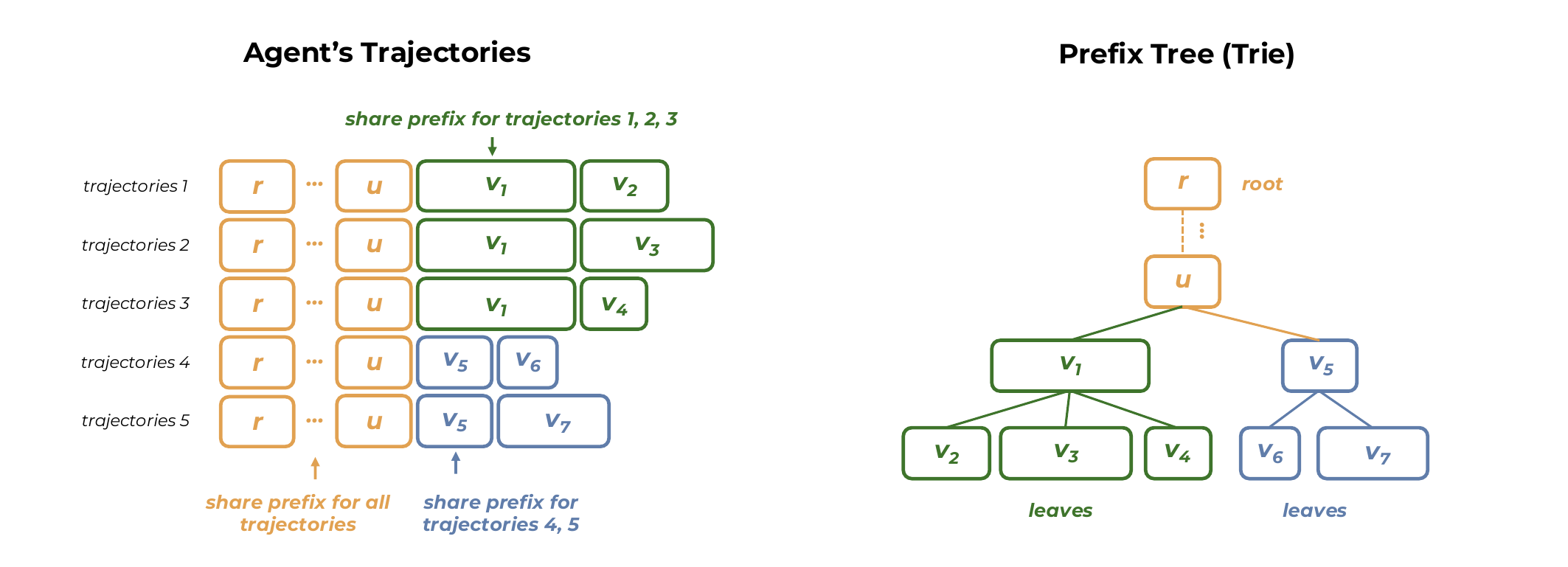}
    \caption{Overview of Tree Training in agentic RL.}
    \label{fig:tree}
\end{figure}

Modern agent scaffolds—particularly those employing sub-agents, concurrent tool invocations, and context engineering—produce training trajectories that are fundamentally tree-structured rather than linear. When a scaffold spawns parallel sub-agents, manages multi-turn context windows with selective retention, or discards intermediate reasoning tokens between turns, the tokens generated by a single task cannot be represented as a flat sequence. The context fed into each subsequent turn is not the direct concatenation of prior turns, yielding branches with deeply shared prefixes. Consequently, naively linearizing these trajectories into independent sequences causes shared prefixes to be recomputed redundantly in every forward and backward pass, imposing a training cost that grows proportionally with the degree of branching. As agent scaffolds grow more sophisticated, this overhead compounds and becomes a fundamental bottleneck.

To eliminate this redundancy, as shown in Figure \ref{fig:tree}, we employ \textbf{Tree Training}~\cite{wang2026treetrainingacceleratingagentic}, which serializes the entire trajectory tree into a single Depth-First Search (DFS) flattened sequence and applies a per-token loss weight. This simple reweighting is provably sufficient: by the linearity of differentiation, the resulting gradients are exactly equivalent to those of the baseline that trains on all root-to-leaf paths independently, requiring only the negligible computational overhead of an element-wise scalar multiplication on the per-token loss tensor. Correct computation further relies on three lightweight components: a tree-structured attention mask (built on FlashAttention V3) that restricts each token's attention to its own root-to-leaf path, per-token position IDs that restore each token's original sequence position rather than its offset in the flattened tree, and the gradient scaling weights described above. The implementation is orthogonal to standard distributed parallelism strategies (TP, EP, DP, PP) and integrates seamlessly with the rest of the training infrastructure. On real-world agentic RL rollouts collected from the diverse scaffolds described above, Tree Training achieves up to a \textbf{6.2$\times$} end-to-end training speedup.

\begin{figure}[tbp]
    \centering
    \includegraphics[
        width=\linewidth,
    ]{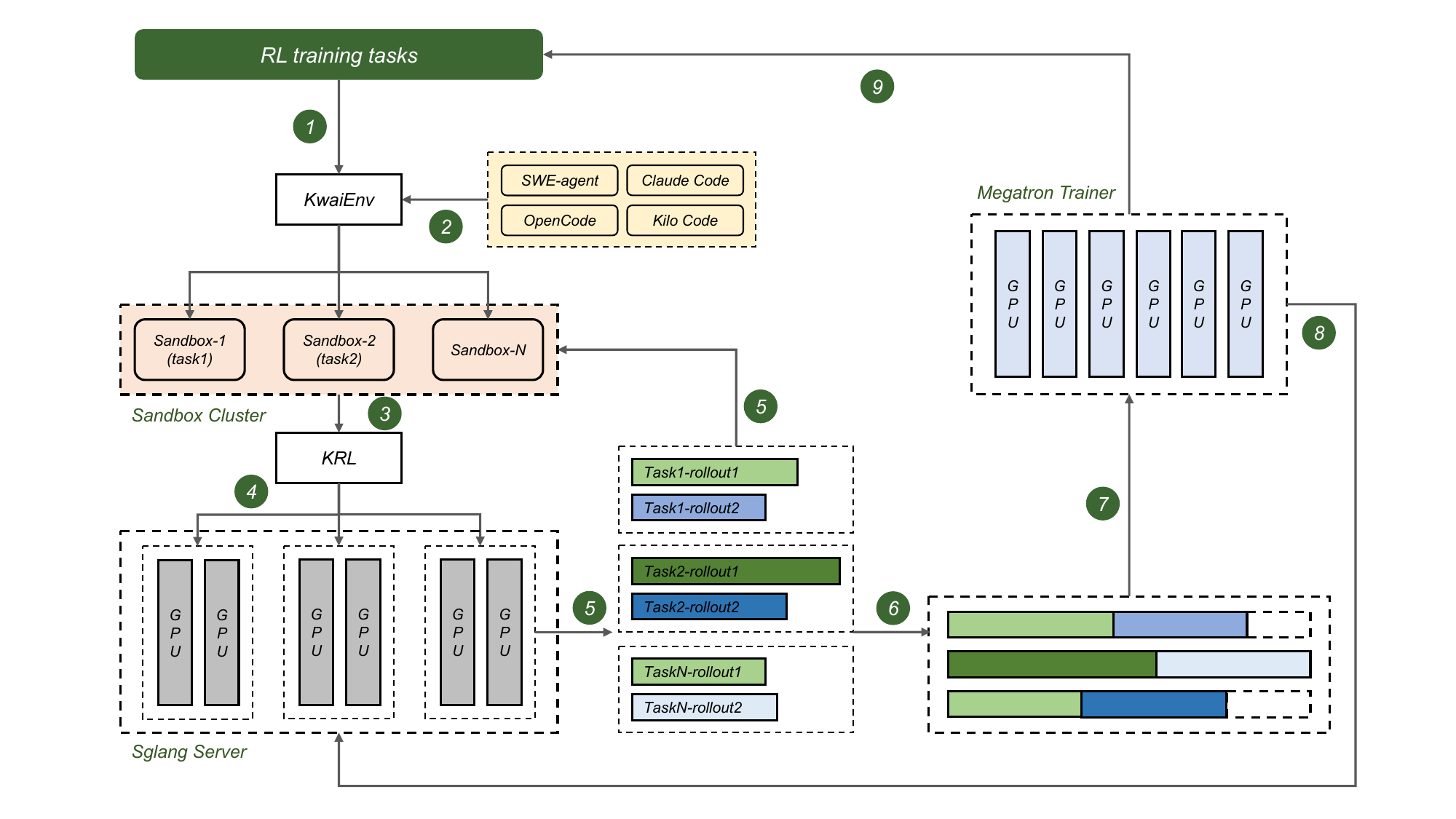}
    \caption{Overview of the RL and sandbox framework.}
    \label{fig:rl}
\end{figure}

\paragraph{High-Concurrency Sandbox Training.}

To efficiently orchestrate the complex interactions between the policy model and diverse execution scaffolds, our framework implements a highly concurrent, asynchronous pipeline. As illustrated in Figure~\ref{fig:rl}, the end-to-end training loop proceeds through the following stages:

\begin{enumerate}
    \item \textbf{Instance Sampling}: We sample a training batch of executable tasks from the dataset.
    \item \textbf{Agent Allocation}: Utilizing KwaiEnv, the system assigns a specific agent scaffold type to each sample and dispatches the execution request to the remote sandbox cluster.
    \item \textbf{Sandbox Initialization}: Powered by \textbf{Wanqing} (Kuaishou's proprietary large-scale container cloud platform), the remote sandbox cluster dynamically provisions the appropriate Docker-isolated environments. Each sandbox securely encapsulates a single execution trajectory and initiates interaction requests to the \texttt{SGLang} inference engine.
    \item \textbf{Request Routing}: The KRL router dynamically orchestrates incoming requests across multiple \texttt{SGLang} inference servers to guarantee strict load balancing.
    \item \textbf{Rollout Generation}: The \texttt{SGLang} inference service iteratively generates trajectory data, totaling a volume of $\text{batch\_size} \times \text{group\_size}$. For agentic multi-turn scenarios, the model actively interacts with the sandbox environment until task completion, maximum turn limits, or time constraints are reached.
    \item \textbf{Reward Computation}: Rollouts are evaluated against task-specific rules or verifier models to acquire environmental rewards. The system calculates advantages and subsequently executes trajectory packing across the rollouts.
    \item \textbf{Engine Switching}: We perform a critical live context switch, transitioning the active GPU resources from hosting the \texttt{SGLang} inference service to the \texttt{Megatron} training service.
    \item \textbf{Parameter Optimization}: The model undergoes policy training and parameter updates using the packed trajectories. Following the update, the newly refined model weights are seamlessly synchronized back to the \texttt{SGLang} servers.
    \item \textbf{Iteration}: The unified cycle instantly advances to train the next batch.
\end{enumerate}

\subsection{Expert Fusion via On-Policy Distillation} 

After developing highly specialized experts across diverse domains (e.g., coding, reasoning), we face the challenge of amalgamating them into a single omni-capable model. Direct weight averaging causes catastrophic forgetting, while standard RL provides feedback too sparse to pinpoint intermediate reasoning errors. Conversely, off-policy SFT suffers from exposure bias. To bridge this gap, we adopt \textbf{On-Policy Distillation (OPD)~\footnote{https://thinkingmachines.ai/blog/on-policy-distillation}}, which combines the active exploration of RL with the dense supervision of knowledge distillation.

During training, the unified Student model actively generates complete trajectories across mixed-domain prompts. \textbf{Crucially, we jointly optimize the Student using both a standard RL loss and an expert-guided OPD loss.} For the RL component, the environment (e.g., execution sandboxes) provides sparse grounding rewards to ensure final task success. Concurrently, for the OPD component, we dynamically select the best-performing expert for each specific task to act as the Teacher. This designated expert evaluates the Student's on-policy rollouts and provides dense, step-level supervision via its \textbf{log-probabilities}. This approach avoids computationally expensive full-logit distillation while providing unbiased optimization targets.

Condensing diverse, specialized capabilities into a single set of model weights inevitably incurs a slight performance degradation compared to the isolated experts, primarily due to capacity constraints and cross-domain interference. However, our joint optimization effectively mitigates catastrophic forgetting. By aligning the Student's active reasoning with the best experts' log-probabilities while grounding it in environmental success, OPD successfully minimizes this performance drop, ultimately yielding a robust and highly capable unified model.

\section{Evaluation}
\label{sec:evaluation}

To systematically evaluate the capabilities of KAT-Coder-V2, we conducted a comprehensive analysis of its performance across multiple representative benchmarks based on the KwaiEnv evaluation platform. This evaluation covers four core dimensions: multi-scaffold coding capability, agent task execution capability, frontend aesthetics generation capability, and general task processing capability. Overall, the results indicate that KAT-Coder-V2 demonstrates outstanding performance across these dimensions, placing it among the top tier of coding models.


\subsection{Multi-Scaffold Coding}
In real-world AI coding scenarios, developers often choose different development scaffolds based on personal habits, team norms, or specific business requirements. These scaffolds often exhibit significant differences in prompt construction, Tool Use protocols, and context organization and management mechanisms, thereby posing high demands on the cross-scaffold generalization capabilities of foundation models.

We evaluated KAT-Coder-V2 across major mainstream scaffolds, using their native interaction protocols and system prompt configurations, on SWE-bench Verified \cite{jimenez2024swebench}, SWE-bench Multilingual \cite{yang2025swe}, and a subset of Swe-rebench-V2 \cite{badertdinov2026swe}. Evaluation results show that the model maintains stable performance across different scaffold environments. Its core metrics on mainstream frameworks such as Claude Code, OpenClaw, and OpenCode are comparable to the most advanced coding model. Thanks to the outstanding framework generalization capability, KAT-Coder-V2 is compatible with over 10 mainstream AI coding scaffolds, providing developers with flexible choices.

\newcolumntype{C}{>{\centering\arraybackslash}X}
\begin{table}[htbp]
    \centering
    \renewcommand{\arraystretch}{1.25}
    \caption{ Evaluation Results on Software Engineering Tasks under Multiple Scaffolds. Data points marked with * are taken from https://www.anthropic.com/news/claude-opus-4-6. Other data points are evaluated on KwaiEnv. }
    \label{tab:exp_multiple_scaffolds}
    \begin{tabularx}{\textwidth}{lCCCCCC}
        \toprule
        \textbf{Benchmark} & \textbf{Scaffold} & \textbf{KAT-Coder-V2} & \textbf{Claude Opus 4.6} \\
        \midrule
        \multirow{3}{*}{SWE-bench Verified}  & Claude Code & 79.6 & 80.8$^{*}$  \\
                                            & OpenCode & 74.8 & 75.0 \\
                                            & OpenClaw & 72.8 & 75.7  \\
        \cline{1-4} 
        \multirow{2}{*}{SWE-bench Multilingual}   & Claude Code & 75.4 & 77.8$^{*}$  \\
                                                & OpenCode & 71.2 & 70.2  \\
        \cline{1-4} 
        \multirow{2}{*}{SWE-rebench-V2 (subset)}   & Claude Code & 43.3 & 43.7  \\
                                                & OpenCode & 38.7 & 37.3  \\                                        
        \bottomrule
    \end{tabularx}
\end{table}


\subsection{Agent Task Execution}
With the rapid rise of new-generation AI Agent frameworks represented by OpenClaw, AI coding tools are further evolving towards the autonomous execution of real-world complex tasks. This trend introduces more complex skill invocation, task orchestration mechanisms, and dynamic scheduling strategies, placing higher-order demands on the model's interaction and tool use capabilities.

To evaluate KAT-Coder-V2 under these conditions, we performed systematic testing on the PinchBench and Claw-Eval benchmarks based on the OpenClaw framework. The results indicate that under stress scenarios such as scheduled triggering, high-concurrency request processing, and long-chain task execution, KAT-Coder-V2 demonstrates strong execution efficiency and response stability.



\newcolumntype{C}{>{\centering\arraybackslash}X}

\begin{table}[htbp]
    \centering
    \renewcommand{\arraystretch}{1.25}
    \setlength{\tabcolsep}{3pt} 
    
    \caption{Evaluation Results on Real-World Agent Benchmarks under OpenClaw. Data points of other models are taken from https://pinchbench.com/ and https://claw-eval.github.io/ (retrieved on March 25, 2026).}
    \label{tab:exp_openclaw}

    \begin{tabularx}{\textwidth}{llCCCCCC}
        \toprule
        \small \textbf{Benchmark} & 
        \small \textbf{Scaffold} & 
        \small \textbf{KAT-Coder-V2} & 
        \small \textbf{GLM-5} & 
        \small \textbf{MiniMax} & 
        \small \textbf{Claude} & 
        \small \textbf{GPT-5.4} & 
        \small \textbf{Gemini} \\[-3ex]
        
        & 
        & 
        \small \textbf{} & 
        & 
        \small \textbf{M2.7} & 
        \small \textbf{Opus 4.6} & 
        & 
        \small \textbf{3.1 Pro} \\
        \midrule
        \multirow{2}{*}{PinchBench}   & Best Score & 88.7 & 86.4 & 87.1 & 87.4 & 90.5 & 86.7   \\
                                      & Average Score & 81.9 & 80.3 & 81.8 & 82.3 & 81.6 & 75.9  \\
        \midrule 
        \multirow{2}{*}{Claw-Eval}   & Pass \^{} 3 & 55.6 & 57.7 & 51.9 & 66.3 & 66.3 & 50.0  \\
                                      & Average Score & 73.4 & 73.0 & 70.7 & 79.3 & 80.6 & 74.2  \\                                        
        \bottomrule
    \end{tabularx}
\end{table}

\subsection{Frontend Aesthetics Generation}
Interface aesthetics are an important component of frontend generation quality, directly affecting users' visual perception and interactive experience. Targeting this dimension, we constructed a systematic aesthetic evaluation benchmark specifically oriented toward reference-free design scenarios. This benchmark covers three typical application scenarios: Landing Pages, Slides, and Data Visualization. Among them, Landing Pages are further divided into 10 evaluation dimensions, while Slides and Data Visualization are each divided into 5 dimensions. Standardized anchor-based scoring scales were designed for each dimension.

All queries in the evaluation test set are based on the colloquial expressions of ordinary users, and the assessments are conducted through blind evaluations by a professional UI/UX designer team under standardized experimental conditions. 

The results show that KAT-Coder-V2 achieved leading aesthetic scores across all three scenarios, demonstrating strong user intent understanding and frontend visual generation capabilities.

\newcolumntype{C}{>{\centering\arraybackslash}X}
\begin{table}[htbp]
    \centering
    \renewcommand{\arraystretch}{1.25}
    \caption{ Evaluation Results on Frontend Generation Task. }
    \label{tab:exp_frontend}
    \begin{tabularx}{\textwidth}{lCCCCCC}
        \toprule
        \textbf{Benchmark} & \textbf{KAT-Coder-V2} & \textbf{GLM-5} & \textbf{Kimi K2.5} \\
        \midrule
        Landing Page & 59.8 & 57.6 & 54.6  \\
        Slides & 57.6 & 42.8 & 34.8  \\
        Data Visualization  & 67.6 & 42.4 & 46.0  \\
        \bottomrule
    \end{tabularx}
\end{table}


\begin{figure}[tbp]
    \centering
    \begin{subfigure}[t]{0.48\linewidth}
        \centering
        \includegraphics[width=\linewidth]{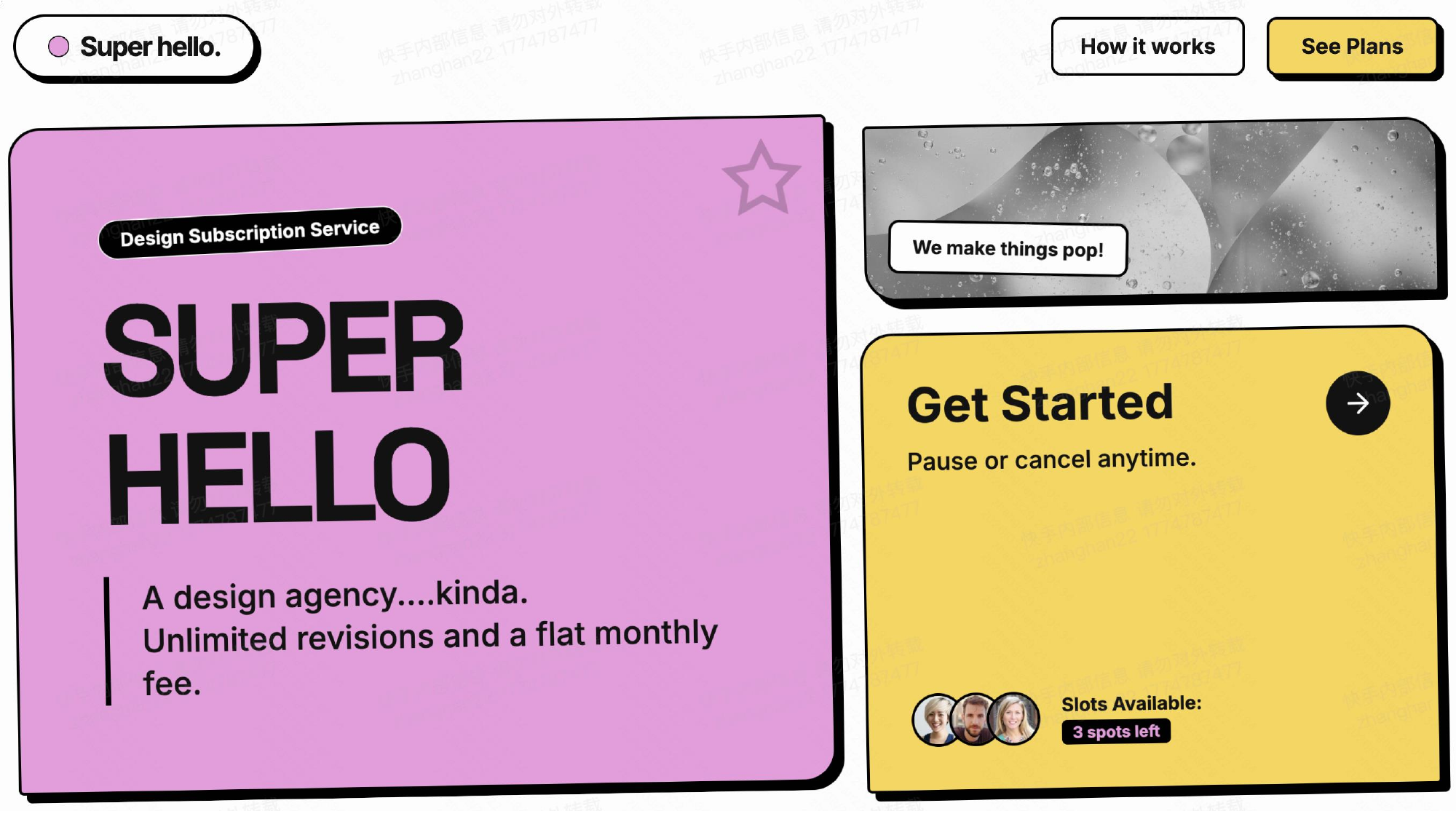}
        \caption{Landing Page 1}
        \label{fig:a}
    \end{subfigure}
    \hfill
    \begin{subfigure}[t]{0.48\linewidth}
        \centering
        \includegraphics[width=\linewidth]{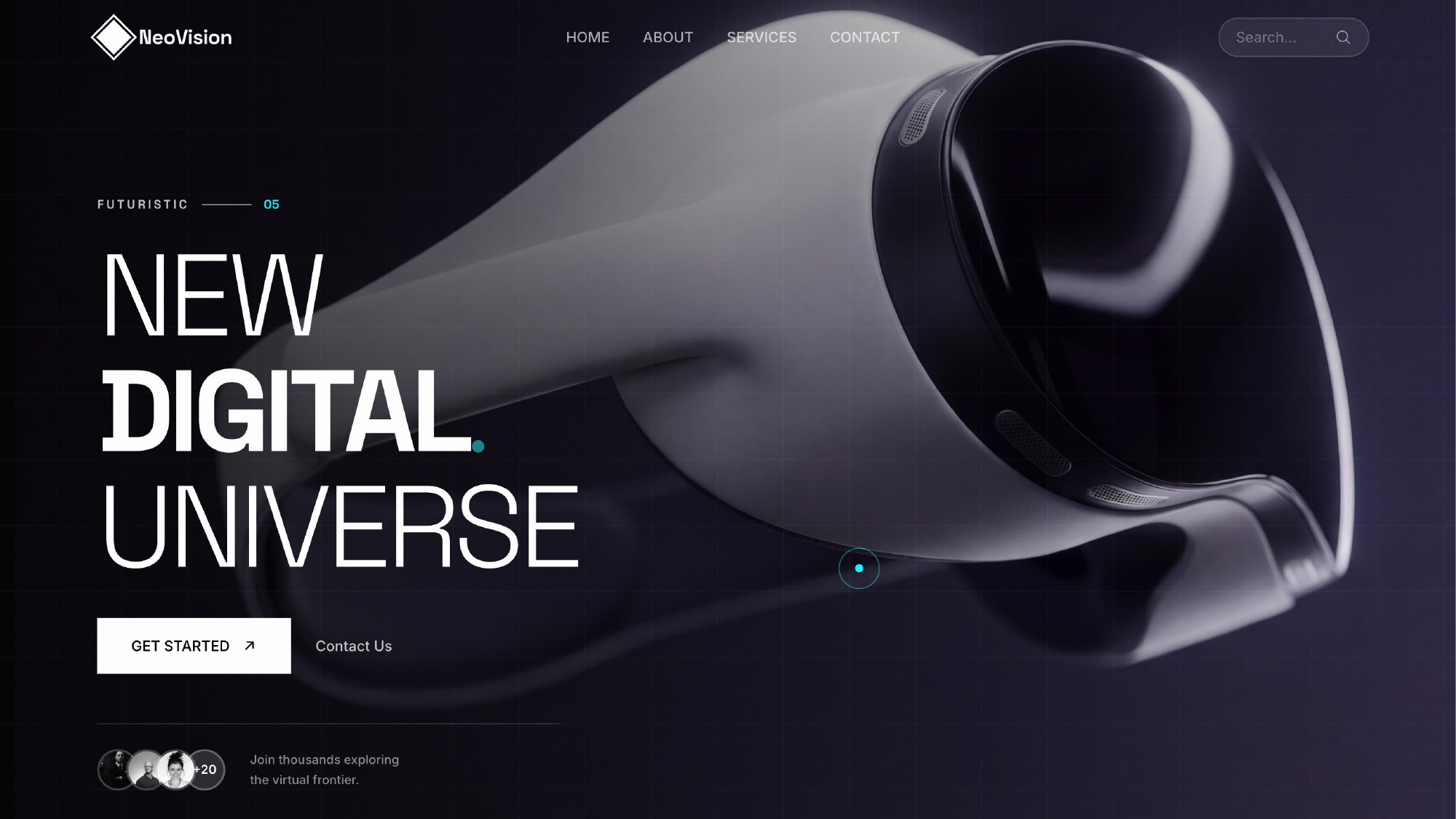}
        \caption{Landing Page 2}
        \label{fig:b}
    \end{subfigure}

    \vspace{0.5em}

    \begin{subfigure}[t]{0.48\linewidth}
        \centering
        \includegraphics[width=\linewidth]{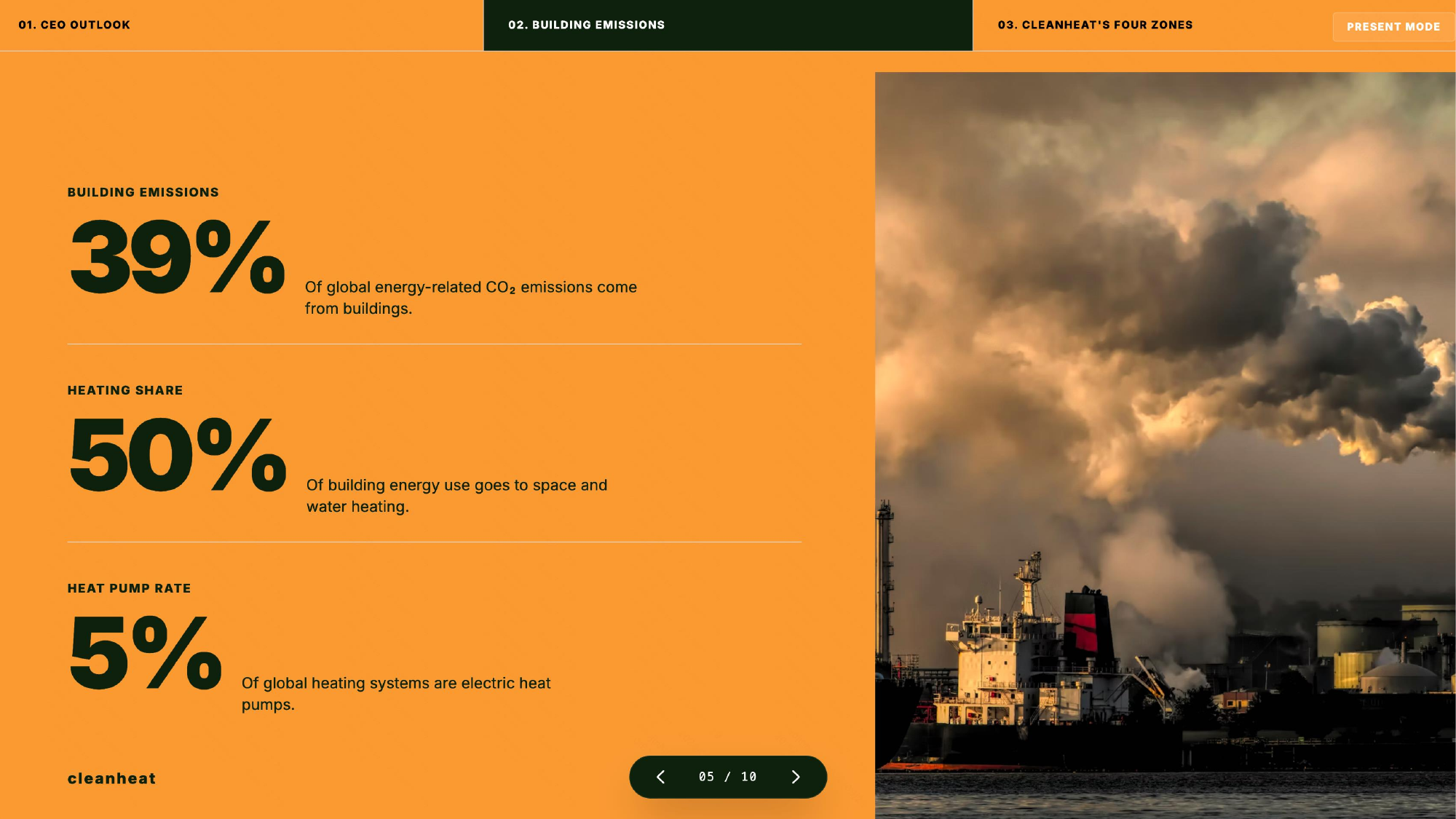}
        \caption{Slides}
        \label{fig:c}
    \end{subfigure}
    \hfill
    \begin{subfigure}[t]{0.48\linewidth}
        \centering
        \includegraphics[width=\linewidth]{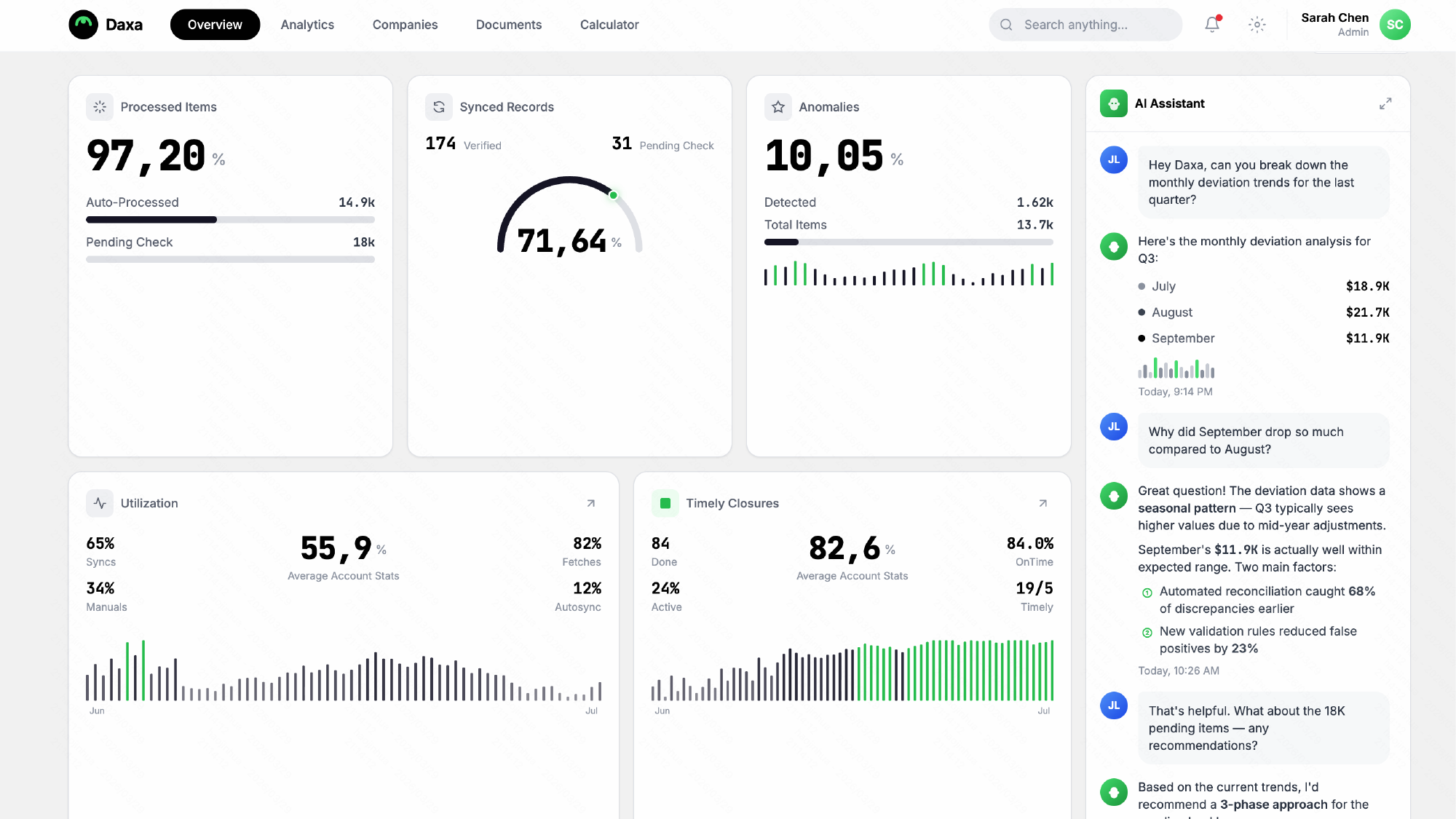}
        \caption{Data Visualization}
        \label{fig:d}
    \end{subfigure}

    \caption[Frontend Generation Results of KAT-Coder-V2]{%
    Frontend Generation Results of KAT-Coder-V2.\protect\\[4pt]
    }
    \label{fig:three_figures}
\end{figure}

\subsection{General Task Processing}
In real-world coding scenarios, a strong model must not only complete basic code generation and optimization but also excel in end-to-end complex tasks, multi-turn interactive reasoning, long-context understanding, and high-precision instruction following.

Focusing on these key capabilities, we systematically evaluated KAT-Coder-V2 on multiple mainstream benchmarks, including Terminal-Bench Hard \cite{merrill2026terminal}, $\tau^2$-Bench Telecom \cite{barres2025tau, yao2024tau}, AA-LCR \cite{artificialanalysis2025lcr}, and IFBench \cite{pyatkin2025generalizing}. The results show that KAT-Coder-V2 achieved competitive scores across various general scenarios. Its strong foundational capabilities provide strong support for its adaptability, robustness, and stability in complex programming environments, further enhancing its comprehensive performance in real-world development tasks.

\newcolumntype{C}{>{\centering\arraybackslash}X}

\begin{table}[htbp]
    \centering
    \renewcommand{\arraystretch}{1.25}
    \setlength{\tabcolsep}{3pt} 
    
    \caption{Evaluation Results on General Task. Data points of KAT-Coder-V2 are evaluated on KwaiEnv. Data points of other models are from https://artificialanalysis.ai/evaluations (retrieved on March 25, 2026).}
    \label{tab:exp_general}
    
    \begin{tabularx}{\textwidth}{lCCCCCC}
        \toprule
        \small \textbf{Benchmark} & 
        \small \textbf{KAT-Coder-V2} & 
        \small \textbf{GLM-5} & 
        \small \textbf{MiniMax} & 
        \small \textbf{Claude} & 
        \small \textbf{GPT-5.4} & 
        \small \textbf{Gemini} \\[-3ex] 
        
        & 
        \small \textbf{} & 
        & 
        \small \textbf{M2.7} & 
        \small \textbf{Opus 4.6} & 
        & 
        \small \textbf{3.1 Pro} \\
        \midrule
        Terminal-Bench Hard & 46.8 & 43.2 & 39.4 & 46.2 & 57.6 & 53.8 \\
        $\tau^2$-Bench Telecom & 93.9 & 98.2 & 84.8 & 92.1 & 91.5 & 95.6 \\
        AA-LCR  & 68.0 & 63.3 & 68.7 & 70.7 & 74.0 & 72.7 \\
        IFBench  & 67.0 & 72.3 & 75.7 & 53.1 & 73.9 & 77.1 \\
        \bottomrule
    \end{tabularx}

\end{table}
\section{Conclusion}
 
In this report, we have introduced KAT-Coder-V2, a comprehensive agentic coding model that demonstrates domain-specialized training, large-scale agentic RL, and unified distillation as a principled path toward building powerful coding agents. By decomposing capabilities into orthogonal expert domains and fusing them through on-policy distillation, KAT-Coder-V2 retains expert-level performance across SWE, frontend generation, terminal reasoning, and general tasks within a single model. The accompanying infrastructure (KwaiEnv) and algorithmic innovations (MCLA, Tree Training), together with systematic agentic scaling across task complexity, prompt diversity, and scaffold generalization, collectively enable stable, efficient training at scale. With these strengths, KAT-Coder-V2 closely rivals the strongest proprietary coding models across multiple scaffolds and benchmarks. However, gaps remain on certain agent execution benchmarks such as Claw-Eval, which we aim to narrow through further scaling of agentic RL and richer environment interaction. Future work will also focus on extending the \textit{Specialize-then-Unify} paradigm to broader agentic domains beyond coding and exploring more efficient expert fusion strategies to fully unlock the potential of domain-specialized training.

\section{Contribution}
\setlength{\parindent}{0pt}
Contributors’ names are listed in alphabetical order by first name. 

\textbf{Core Contributors}

\begin{tabular}{@{}L{0.18\textwidth}@{\hspace{1em}}
                L{0.18\textwidth}@{\hspace{1em}}
                L{0.18\textwidth}@{\hspace{1em}}
                L{0.18\textwidth}@{\hspace{1em}}
                L{0.18\textwidth}@{}}
Fengxiang Li   & Han Zhang      & Haoyang Huang   & Jinghui Wang   & Jinhua Hao \\
Kun Yuan       & Mengtong Li    & Minglei Zhang   & Pengcheng Xu   & Wenhao Zhuang \\
Yizhen Shao    & Zongxian Feng  &                 &                & \\
\end{tabular}

\vspace{0.8em}
\textbf{Contributors}

\begin{tabular}{@{}L{0.18\textwidth}@{\hspace{1em}}
                L{0.18\textwidth}@{\hspace{1em}}
                L{0.18\textwidth}@{\hspace{1em}}
                L{0.18\textwidth}@{\hspace{1em}}
                L{0.18\textwidth}@{}}
Can Tang       & Chao Wang      & Chengxiao Tong  & Fan Yang       & Gang Xiong \\
Haixuan Gao    & Han Gao        & Hao Wang        & Haochen Liu    & Hongliang Sun \\
Jiabao Li      & Jingwen Chang  & Jun Du          & Junyi Peng     & Leizhen Cui \\
Meimei Jing    & Mingqi Wu      & Shangpeng Yan   & Shaotong Qi    & Suzhe Xu \\
Wenxuan Zhao   & Xianda Sun     & Xuan Xie        & Yanbo Wang     & Yao Xia \\
Yinghan Cui    & Yingpeng Chen  & Yong Wang       & Yuze Shi       & Zhiwei Shen \\
Ziyu Wang      &                &                 &                & \\
\end{tabular}

\vspace{0.8em}
\textbf{Tech Leads}

\begin{tabular}{@{}L{0.18\textwidth}@{\hspace{1em}}
                L{0.18\textwidth}@{\hspace{1em}}
                L{0.18\textwidth}@{\hspace{1em}}
                L{0.18\textwidth}@{\hspace{1em}}
                L{0.18\textwidth}@{}}
Ming Sun       & Lin Ye         & Bin Chen        &                & \\
\end{tabular}

\newpage
\bibliography{references}

\begin{thebibliography}{10}

\bibitem{claudeopus46}
{Anthropic}.
\newblock Claude opus 4.6 system card, 2026.

\bibitem{gemini3pro2025}
{Google DeepMind}.
\newblock Gemini 3 pro model card, 2025.

\bibitem{zeng2026glm}
Aohan Zeng, Xin Lv, Zhenyu Hou, Zhengxiao Du, Qinkai Zheng, Bin Chen, Da~Yin, Chendi Ge, Chengxing Xie, Cunxiang Wang, et~al.
\newblock Glm-5: from vibe coding to agentic engineering.
\newblock {\em arXiv preprint arXiv:2602.15763}, 2026.

\bibitem{team2026kimi}
Kimi Team, Tongtong Bai, Yifan Bai, Yiping Bao, SH~Cai, Yuan Cao, Y~Charles, HS~Che, Cheng Chen, Guanduo Chen, et~al.
\newblock Kimi k2. 5: Visual agentic intelligence.
\newblock {\em arXiv preprint arXiv:2602.02276}, 2026.

\bibitem{huang2026step}
Ailin Huang, Ang Li, Aobo Kong, Bin Wang, Binxing Jiao, Bo~Dong, Bojun Wang, Boyu Chen, Brian Li, Buyun Ma, et~al.
\newblock Step 3.5 flash: Open frontier-level intelligence with 11b active parameters.
\newblock {\em arXiv preprint arXiv:2602.10604}, 2026.

\bibitem{liu2025deepseek}
Aixin Liu, Aoxue Mei, Bangcai Lin, Bing Xue, Bingxuan Wang, Bingzheng Xu, Bochao Wu, Bowei Zhang, Chaofan Lin, Chen Dong, et~al.
\newblock Deepseek-v3. 2: Pushing the frontier of open large language models.
\newblock {\em arXiv preprint arXiv:2512.02556}, 2025.

\bibitem{jimenez2024swebench}
Carlos~E Jimenez, John Yang, Alexander Wettig, Shunyu Yao, Kexin Pei, Ofir Press, and Karthik~R Narasimhan.
\newblock {SWE}-bench: Can language models resolve real-world github issues?
\newblock In {\em The Twelfth International Conference on Learning Representations}, 2024.

\bibitem{merrill2026terminal}
Mike~A Merrill, Alexander~G Shaw, Nicholas Carlini, Boxuan Li, Harsh Raj, Ivan Bercovich, Lin Shi, Jeong~Yeon Shin, Thomas Walshe, E~Kelly Buchanan, et~al.
\newblock Terminal-bench: Benchmarking agents on hard, realistic tasks in command line interfaces.
\newblock {\em arXiv preprint arXiv:2601.11868}, 2026.

\bibitem{barres2025tau}
Victor Barres, Honghua Dong, Soham Ray, Xujie Si, and Karthik Narasimhan.
\newblock $\tau^2$-bench: Evaluating conversational agents in a dual-control environment.
\newblock {\em arXiv preprint arXiv:2506.07982}, 2025.

\bibitem{zhan2025kat}
Zizheng Zhan, Ken Deng, Jinghui Wang, Xiaojiang Zhang, Huaixi Tang, Minglei Zhang, Zhiyi Lai, Haoyang Huang, Wen Xiang, Kun Wu, et~al.
\newblock Kat-coder technical report.
\newblock {\em arXiv preprint arXiv:2510.18779}, 2025.

\bibitem{deng2025swebenchproaiagents}
Xiang Deng, Jeff Da, Edwin Pan, Yannis~Yiming He, Charles Ide, Kanak Garg, Niklas Lauffer, Andrew Park, Nitin Pasari, Chetan Rane, et~al.
\newblock Swe-bench pro: Can ai agents solve long-horizon software engineering tasks?
\newblock {\em arXiv preprint arXiv:2509.16941}, 2025.

\bibitem{jain2024livecodebenchholisticcontaminationfree}
Naman Jain, King Han, Alex Gu, Wen-Ding Li, Fanjia Yan, Tianjun Zhang, Sida Wang, Armando Solar-Lezama, Koushik Sen, and Ion Stoica.
\newblock Livecodebench: Holistic and contamination free evaluation of large language models for code.
\newblock {\em arXiv preprint arXiv:2403.07974}, 2024.

\bibitem{zhou2023webarena}
Shuyan Zhou, Frank~F Xu, Hao Zhu, Xuhui Zhou, Robert Lo, Abishek Sridhar, Xianyi Cheng, Tianyue Ou, Yonatan Bisk, Daniel Fried, et~al.
\newblock Webarena: A realistic web environment for building autonomous agents.
\newblock {\em arXiv preprint arXiv:2307.13854}, 2023.

\bibitem{si2025design2code}
Chenglei Si, Yanzhe Zhang, Ryan Li, Zhengyuan Yang, Ruibo Liu, and Diyi Yang.
\newblock Design2code: Benchmarking multimodal code generation for automated front-end engineering.
\newblock In {\em Proceedings of the 2025 Conference of the Nations of the Americas Chapter of the Association for Computational Linguistics: Human Language Technologies (Volume 1: Long Papers)}, pages 3956--3974, 2025.

\bibitem{lin2026cligym}
Yusong Lin, Haiyang Wang, Shuzhe Wu, Lue Fan, Feiyang Pan, Sanyuan Zhao, and Dandan Tu.
\newblock Cli-gym: Scalable cli task generation via agentic environment inversion.
\newblock {\em arXiv preprint arXiv:2602.10999}, 2026.

\bibitem{zhu2026termigen}
Kaijie Zhu, Yuzhou Nie, Yijiang Li, Yiming Huang, Jialian Wu, Jiang Liu, Ximeng Sun, Zhenfei Yin, Lun Wang, Zicheng Liu, Emad Barsoum, William~Yang Wang, and Wenbo Guo.
\newblock Termigen: High-fidelity environment and robust trajectory synthesis for terminal agents.
\newblock {\em arXiv preprint arXiv:2602.07274}, 2026.

\bibitem{shao2024deepseekmath}
Zhihong Shao, Peiyi Wang, Qihao Zhu, Runxin Xu, Junxiao Song, Xiao Bi, Haowei Zhang, Mingchuan Zhang, YK~Li, Yang Wu, et~al.
\newblock Deepseekmath: Pushing the limits of mathematical reasoning in open language models.
\newblock {\em arXiv preprint arXiv:2402.03300}, 2024.

\bibitem{zheng2025group}
Chujie Zheng, Shixuan Liu, Mingze Li, Xiong-Hui Chen, Bowen Yu, Chang Gao, Kai Dang, Yuqiong Liu, Rui Men, An~Yang, et~al.
\newblock Group sequence policy optimization.
\newblock {\em arXiv preprint arXiv:2507.18071}, 2025.

\bibitem{wang2026treetrainingacceleratingagentic}
Shaojie Wang, Jinghui Wang, Yinghan Cui, Xuxing Chen, Chao Wang, Liang Huang, Xiaojiang Zhang, Junyi Peng, Li~Wan, Haotian Zhang, et~al.
\newblock Tree training: Accelerating agentic llms training via shared prefix reuse.
\newblock {\em arXiv preprint arXiv:2511.00413}, 2025.

\bibitem{yang2025swe}
John Yang, Kilian Lieret, Carlos~E Jimenez, Alexander Wettig, Kabir Khandpur, Yanzhe Zhang, Binyuan Hui, Ofir Press, Ludwig Schmidt, and Diyi Yang.
\newblock Swe-smith: Scaling data for software engineering agents.
\newblock {\em arXiv preprint arXiv:2504.21798}, 2025.

\bibitem{badertdinov2026swe}
Ibragim Badertdinov, Maksim Nekrashevich, Anton Shevtsov, and Alexander Golubev.
\newblock Swe-rebench v2: Language-agnostic swe task collection at scale.
\newblock {\em arXiv preprint arXiv:2602.23866}, 2026.

\bibitem{yao2024tau}
Shunyu Yao, Noah Shinn, Pedram Razavi, and Karthik Narasimhan.
\newblock $\tau$-bench: A benchmark for tool-agent-user interaction in real-world domains.
\newblock {\em arXiv preprint arXiv:2406.12045}, 2024.

\bibitem{artificialanalysis2025lcr}
Artificial~Analysis Team.
\newblock Artificial analysis long context reasoning benchmark(lcr), 2025.

\bibitem{pyatkin2025generalizing}
Valentina Pyatkin, Saumya Malik, Victoria Graf, Hamish Ivison, Shengyi Huang, Pradeep Dasigi, Nathan Lambert, and Hannaneh Hajishirzi.
\newblock Generalizing verifiable instruction following.
\newblock In {\em The Thirty-ninth Annual Conference on Neural Information Processing Systems Datasets and Benchmarks Track}, 2025.

\end{thebibliography}

\end{document}